\documentclass{article}

\usepackage{PRIMEarxiv}

\usepackage[utf8]{inputenc} 
\usepackage[T1]{fontenc}    
\usepackage{hyperref}       
\usepackage{url}            
\usepackage{booktabs}       
\usepackage{amsfonts}       
\usepackage{nicefrac}       
\usepackage{microtype}      
\usepackage{lipsum}
\usepackage{fancyhdr}       
\usepackage{graphicx}       
\graphicspath{{./images/} }    

\title{Explore the Expression: Facial Expression Generation using Auxiliary Classifier Generative Adversarial Network
}

\author{
  J.Rafid Siddiqui \\
  \\
  AZAD Research Lab \\
  Sweden\\
  \texttt{jrs@azaditech.com} 
}

\begin{document}
\maketitle

\begin{abstract}
Facial expressions are a form of non-verbal communication that humans perform seamlessly for meaningful transfer of information. Most of the literature addresses the facial expression recognition aspect however, with the advent of Generative Models, it has become possible to explore the affect space in addition to mere classification of a set of expressions. In this article, we propose a generative model architecture which robustly generates a set of facial expressions for multiple character identities and explores the possibilities of generating complex expressions by combining the simple ones.
 
\end{abstract}


\section{Introduction}

The ability to communicate effectively is perhaps what makes humans worthy of the title 'social animal'. It is not only the prowess that is noteworthy but the diverseness of the forms of communication that humans have devised in order to convey information. Humans use a range of tools to convey the content of one brain to another brain; speech, handwritten/typed text, visual content(e.g. shapes,drawings,illustrations,images,videos etc.), facial expressions, and body language to name a few. The list of tools keeps on expanding with the advancement in the modern technology. However, there are some basic tools that humans use readily in order to accomplish the task of effective communication (specially the social communication). Facial expressions are such a tool that is built into human's subconscious development.It is not common for most species to have an extended childhood like human babies do; human babies spend multiple years dedicated for brain's development with limited to no physical movement. The majority of the brain's input in these years consists of facial images, mostly familiar faces with expressions. This leads to a lasting imprint on the subconscious and therefore, the face recognition and facial expressions as a mean of effective communication becomes the primary and instinctive mode of human communication. \\

This importance of facial information is evident from the amount of research that has been dedicated to mere face recognition over the past hundred years. With the success of deep learning models in the recent years, it is now a task that can be performed by machines effectively as well. \cite{parkhi_deep_2015}. However, facial expression research is still an on-going research area, with some success over facial expression recognition tasks in the recent years. \\

In addition to Facial expression recognition, facial expression generation is also an important task whereby the aim is not just to classify an image into a set of class labels but to generate an image exhibiting a particular facial expression. This is not only more challenging task but it is more interesting as it involves modeling the underlying mechanics of the facial expression and therefore, achieve not just the mere labeling system but a system that can understand the expressions on a deeper level. This opens up many application areas such as photo/video editing and character animations to name a few. \\

Generative Adversarial Network (GAN) is a convolutional neural network architecture that was first proposed by Ian Goodfellow in 2014 \cite{goodfellow_generative_2014}. Since then it has generated a plethora of different variations and advancements and solved many problems in computer vision. A simple GAN architecture consists of a generator and a discriminator network both of which work antagonistically and yet achieve the common goal. A generator network is a network that takes a random sample drawn from Gaussian distribution, termed as a latent vector and constructs a desired image by minimizing the loss function. The discriminator network takes in the real image from the training set and the fake output of the generator network and classifies it as either a real or a fake image. A variation in the GAN architecture was later proposed which utilized Convolutional layers in addition to dense network and was termed Deep Convolutional GAN (DCGAN)\cite{radford_unsupervised_2016}. The original GAN architecture did not have any way of telling the generator to produce a certain type of image, however, a conditional GAN (cGAN) which takes a prior in the form of a class label can produce images of a particular class \cite{mirza_conditional_2014}. Although cGAN has a generator architecture that can handle multiple class labels but discriminator still produces a binary output. Since most of the real world data consists of datasets with multiple classes, a variant called Auxiliary Classifier (AC-GAN) was introduced which extends cGAN by introducing a discriminator network that outputs not only the binary decision but also the class labels \cite{odena_conditional_2017}. In this article we shall build upon the concept of AC-GAN and propose an architecture that robustly solves the facial expression generation problem.

\section{Related Work}
\label{sec:related-work}

As image generation is relatively new field of research, most of the closest work exist in the domain of facial expression recognition. Facial expression recognition has been studied for many decades but the the most relevant work would be which was published in the recent years and has used deep learning method in their solution. In this regard, a few works are noteworthy to mention here. In \cite{kuo_compact_2018} a robust frame based facial expression recognition method is proposed which classifies the input images into a set of affect classes over multiple dataset of face images. In \cite{khaireddin_facial_2021} authors use a CNN architecture (VGGNet)and achieve state-of-the-art accuracy (at the time of the writing) over FER2013 dataset by finetuning the parameters. Another similar CNN based approach was reported in \cite{sajjanhar_deep_2018} and evaluated the model with pre-trained face recognition models\cite{cao_vggface2_2018}. A more GAN related work was reported in  \cite{ning_emotiongan_2020} in which authors proposed a method which encodes the given image into feature space and then uses a Generalized Linear Model (GLM) to fit the general direction of different facial expressions in the feature space. Another GAN based work was proposed in \cite{deng_cgan_2019} which uses one generator and three discriminators based architecture and could transform an input image along with an affect condition to another emotion. A similar GAN based architecture was reported in \cite{ali_facial_2019} which uses disentangled GAN network in order to accomplish the same task. 

\section{Method}
\label{sec:method}

In this section we detail the proposed Facial Expression GAN (FExGAN)architecture, construction of cost function, data preparation and the training setup used for experiments. 

\subsection{FExGAN Architecture}

\begin{figure}[!htb]
  \centering
  \includegraphics[scale=.5]{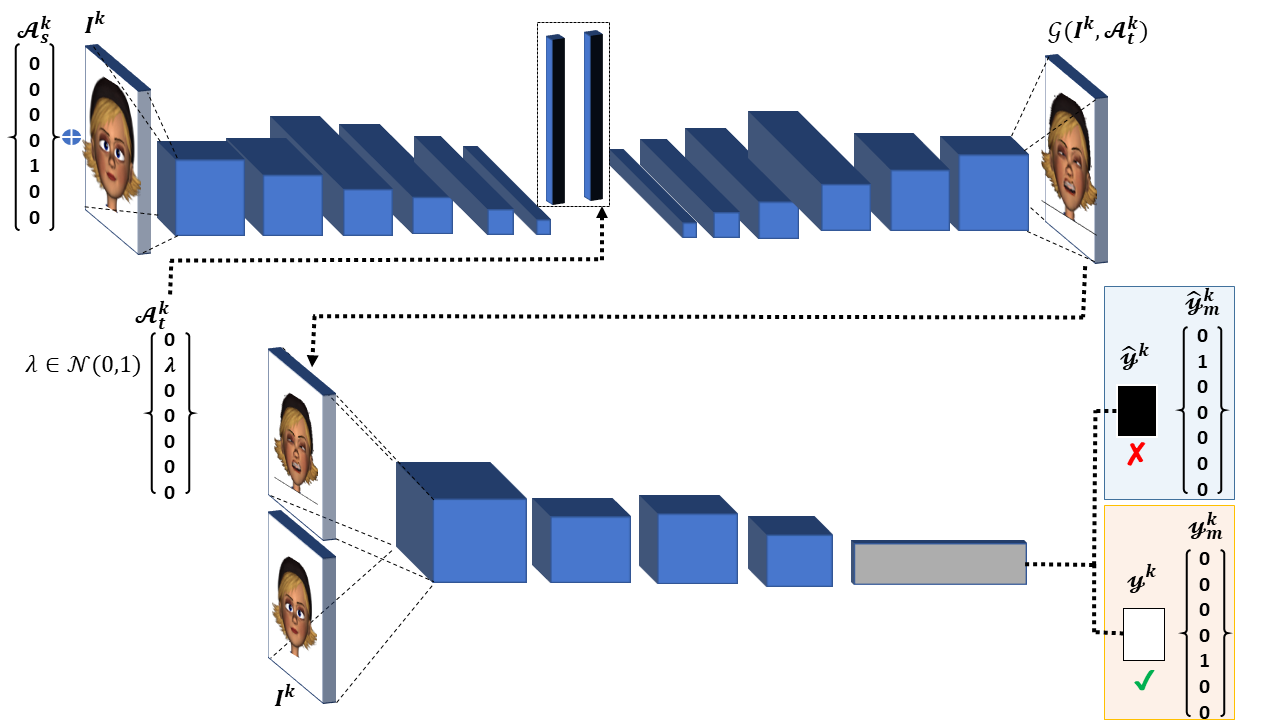}
  \caption{FExGAN Architecture}
  \label{fig:fig1}
\end{figure}

The generator in a classical GAN architecture consists of only up-sampling layers whereby an input Gaussian vector is transformed into a fake output image, however, the reconstruction performance of a generator can be significantly improved if the latent vector comes from the same distribution as of the samples in the dataset. The images of a particular identity in the dataset form a similarity cluster, therefore, the image of each character identity $\mathcal{I}^k$ in a latent space $\mathcal{Z}(\mathcal{I}^k) , \mathcal{Z} \subset \mathcal{N}(\mu^n,\sigma^n)$ can be approximated to be sampled from a normal distribution. One way to embed into such a distribution in latent space is by using an encoder in addition to the downsampling layers in the generator architecture. We first add an encoder section which embeds the input image into a latent distribution sampled from normal distribution and then decode the fake image out from the embedded space. The architecture of such encoding and decoding network can be seen in figure \ref{fig:fig1}. \\

\par
 
Since we want to not only regenerate an image similar to an input image but also be able to manipulate the facial expression, we input a condition (i.e. an affect) in addition to the input image to the encoder and decoder as well. The condition $\mathcal{A}_s^k=[0,0,0,0,1,0,0]$ is a 1-dimensional vector with elements equal to the number of class labels in the dataset. The constraint on latent vector to be from a normal distribution allows us to model the variations within the images of a certain character, however, we would also like to manipulate the variations that exists within each affect. A typical sparse binary vector would be sufficient for classification but in order to model the variations,we sample the target affect $\mathcal{A}_t^k(\lambda) , \lambda \subset \mathcal{N}(0,1)$ from the normal distribution as well. We pass the sampled condition vector(i.e. an affect) to the decoder for each image in the trainingset. This allows us to modify the affect while keeping the character identity the same when training the model. \\

\par
The output of the generator network feeds into the discriminator network which is responsible for determining the validity of the generated output and distinguishing between real and fake outputs. In contrast to a typical conditional GAN, we output not only the decision about real or a fake output but also output a classification label of the input image.  This auxiliary classification architecture of the discriminator network allows us to train on a multi-class dataset which is usually the case in the real world applications. \\

\subsection{Generator Architecture}

\begin{figure}[!htb]
  \centering
  \includegraphics[scale=.35]{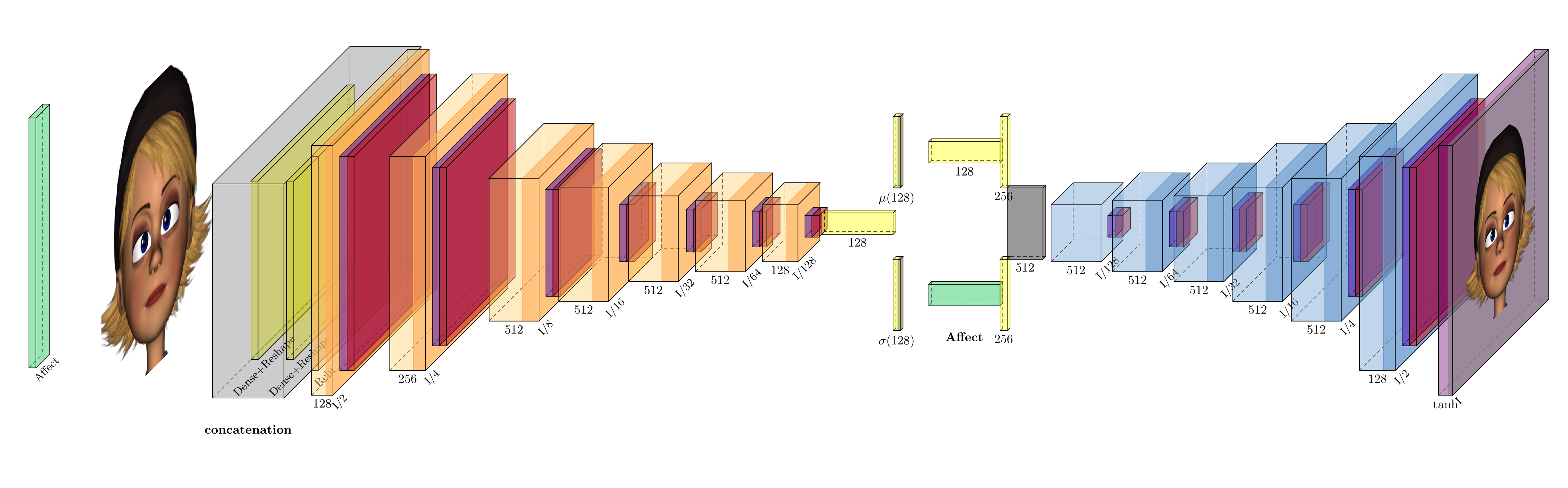}
  \caption{Layers in the Generator Network}
  \label{fig:fig2}
\end{figure}

As explained in the previous section, the generator consists of two major networks: an encoder and a decoder network. An encoder network is responsible for projecting the input image given the prior condition, to a latent space distribution $\mathcal{Z}(\mathcal{I}^k) , \mathcal{Z} \subset \mathcal{N}(\mu^n,\sigma^n)$. The construction of encoder consists of multiple blocks of downsampling layers. Each block has three operations: Convolutions with strides, batch normalization and activation function output. The 2D convolutions with strides allows the convolutions and downsampling in one step which can also be considered as an approximation of a separate convolution and a pooling step often performed in some network architectures. The Batch-Normalization is done by normalizing each batch with its respective mean and variance. The activation layer is a Relu layer which keeps the outputs within a positive range. The downsampling blocks of layers are repeated multiple times, dividing the image size in half at each block output, until the image becomes uni-dimensional. The features learned at each block are increased in the first two layers and then kept to a fixed number consistently in each layer block. The output of the last downsampling block is flattened and is connected to two layers, $\mu$ and $\sigma$ layers with n dimensions. These layers represent the normal distribution in the latent space. A latent layer is then added as an output of the encoder network which merges the outputs of $\mu$ and $\sigma$ layers and results a latent vector by sampling randomly from normal distribution. The layer blocks in the encoder network can be seen in the figure \ref{fig:fig2}. \\

The decoder network takes the output of the encoder as well as a condition vector $\mathcal{A}_t^k(\lambda) , \lambda \subset \mathcal{N}(0,1)$ (i.e. the desired affect). Each input is connected to a dense layer(s) and then concatenated and reshaped to form a 2D vector, which is fed into a set of upsampling blocks. Each upsampling layer block consists of three layers: a conv2D transpose layer, a Batch-Normalization layer and an activation function. The conv2D transpose with strides achieves the opposite of conv2D with strides. The activation function used in the decoder network is LeakyRelu as according to the best practices as evident from the prior research in the field. There are as many upsampling block as there are the downsampling blocks. The last layer of the decoder network uses a tanh as an activation function in order to facilitate the reconstruction of an image. \\

\subsection{Discriminator Architecture}

\begin{figure}
  \centering
  \includegraphics[scale=.5]{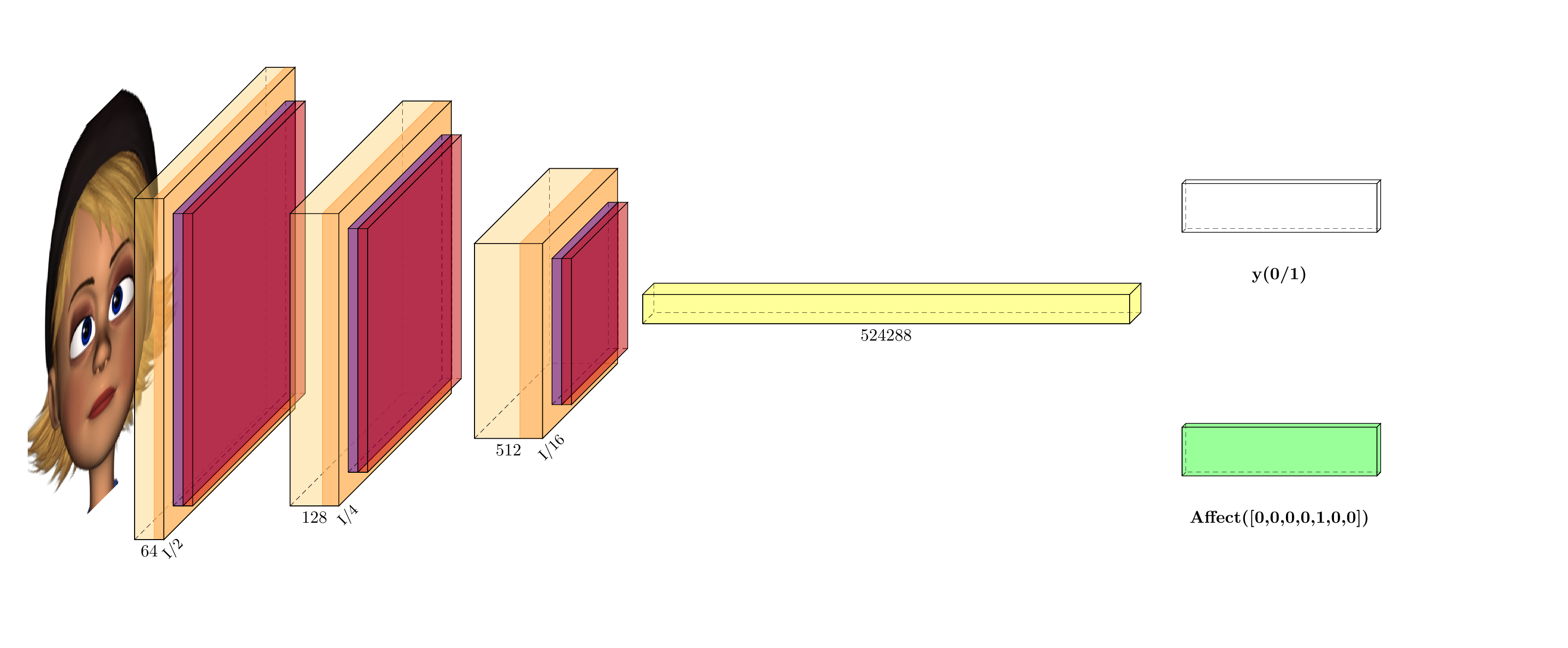}
  \caption{Layers in the Discriminator Network}
  \label{fig:fig3}
\end{figure}

The discriminator network consists of a simple network that takes an input image and ouputs the decision of whether it is a real/fake along with the class it belongs. There are three major layer blocks in the discriminator network. Each layer block consists of three layers: a 2D convolution layer with strides, a Batch-Normalization layer, and a Relu activation function. The features learned at each block are increased steadily and then flattened for a final dense classification network. The final set of layers are a sigmoid layer responsible for real/fake decision while a softmax layer, being responsible for prediction of facial expression of the input image. The discriminator network can be seen in figure\ref{fig:fig3}.

\subsection{Cost Function}

A GAN architecture functions in a mini-max fashion whereby the generator network tries to maximize the loss function while the discriminator tries to minimize it. Since we have an encoder and a decoder network blocks in the generator, we shall include the reconstruction loss in addition to the GAN loss. The GAN loss in the generator network for $\mathcal{I}^k$ is written in equation \ref{eq:eq1}.

\begin{equation}
\label{eq:eq1}
\mathcal{L}^\mathcal{G}_{GAN}(\mathcal{I}_g^k,\mathcal{A}_t^k,\hat{y}^k,\hat{y}_m^k) = \phi_b(1,\hat{y}^k) + \phi_m(\mathcal{A}_t^k,\hat{y}_m^k)
\end{equation}

where $\hat{y}^k,\hat{y}_m^k = \mathcal{D}(\mathcal{I}_g^k)$, $\phi$ and $\phi_m$ are outputs of the discriminator $\mathcal{D}$ on fake image $\mathcal{I}^k_g=\mathcal{G}(\mathcal{I}^k_s,\mathcal{A}_t^k )$ generated by generator $\mathcal{G}$, binary cross entropy and multi-class entropy functions respectively. \\

The generator tries to not just reproduce the input image but rather it tries to produce an output that satisfies the given target affect $\mathcal{A}_t^k$ therefore, the reconstruction should minimize the difference between the generated image $\mathcal{I}^k_g$ and the target image $\mathcal{I}^k_t$ which is randomly selected from the list of images of the desired affect while keeping the identity of the character the same. The reconstruction loss therefore, is an L1-loss between the generated and the target image and is given in equation \ref{eq:eq2}. \\

\begin{equation}
\label{eq:eq2}
\mathcal{L}_{reconst}(\mathcal{I}^k_s,\mathcal{I}^k_t,\mathcal{A}_t^k) = \sum|\mathcal{G}(\mathcal{I}^k_s,\mathcal{A}_t^k )-\mathcal{I}^k_t|
\end{equation}

In earlier section, we discussed that the generated image comes from a normal distribution in the latent space, therefore, we shall enforce a constraint on the generated latent vector by introducing a KL-divergence loss. The KL-loss on the output of the decoder is given in equation \ref{eq:eq3}.

\begin{equation}
\label{eq:eq3}
\mathcal{L}_{KL}(\mu^n_i,\sigma^n_i) = \frac{-1}{2} \sum_i(1+\sigma^n_i -(\mu^n_i)^2 - e^{\sigma^n_i} ) 
\end{equation}

The total loss of the generator then be obtained by combining the GAN loss, reconstruction loss and the KL-loss and is given in the equation \ref{eq:eq4}.

\begin{equation}
\label{eq:eq4}
\mathcal{L}_{gen} = \alpha \mathcal{L}^\mathcal{G}_{GAN} + \beta \mathcal{L}_{KL} + \gamma \mathcal{L}_{reconstruction} 
\end{equation}

Due to the adversarial nature of the construction, the discriminator tries to minimize the GAN loss by computing the binary and multi-class entropies for real as well as for the fake/generated images. The discriminator loss for the real and the fake images is given in equation \ref{eq:eq5} and \ref{eq:eq6} respectively.

\begin{equation}
\label{eq:eq5}
\mathcal{L}_{real}(\mathcal{A}_s^k,y^k,y_m^k) = \phi_b(1,y^k) + \phi_m(\mathcal{A}_s^k,y_m^k)
\end{equation}

\begin{equation}
\label{eq:eq6}
\mathcal{L}_{fake}(\mathcal{A}_t^k,\hat{y}^k,\hat{y}_m^k) = \phi_b(0,\hat{y}^k) + \phi_m(\mathcal{A}_t^k,\hat{y}_m^k)
\end{equation}

The discriminator loss then becomes the combined loss on the real and fake images and is given in equation \ref{eq:eq7}.

\begin{equation}
\label{eq:eq7}
\mathcal{L}^\mathcal{D}_{GAN} = \mathcal{L}_{real} + \mathcal{L}_{fake}
\end{equation}

\section{Experimentation}

In this section, we describe the process of data preparation, model training and experimentation. \\

\subsection{Data Preparation}

The dataset we have used in this study is a synthetic images of cartoon characters and has been created for the purpose of facial expression research in an earlier work \cite{aneja_modeling_2017}. The dataset consists of 55.76K images of six cartoon characters with seven different affects. Each affect has ~10K images with various degrees of facial expressions for a particular affect. This provides a rich source of information to learn from, which is hard to obtain when using images of real humans. It is also interesting because facial expression transfer and manipulation is more applicable to a character animation setting rather than on static images of real humans. \\
we pre-process each input image in the dataset with a set of operations. In particular, we first resize the image, normalize and then add random noise by cropping and adding jitter. The normalization is a standard normalization which makes the images between -1 and 1 range. Once all the images are pre-processed, we split the dataset into training and a validation set. The training set consists of 44.6K images while the validation set consists of 11.1K respectively in this study. \\

\subsection{Model Training}

The training-set obtained as a result of pre-processing from the previous section is used to train the generator and discriminator model separately. However, since the loss is constructed in an adversarial manner, both the generator and the discriminator behave in adversarial fashion. In each training step, loss for both the generator and the discriminator is computed as according to the earlier section and weights are updated accordingly. The learning-rate of \emph{2e-4} has been used with \emph{adam} as an optimization algorithm. The target images for each batch, are selected randomly for the character in the source image and according to the desired affect. The training is performed in batches with a batch size of 32 in this study and process is iterated for 100 epochs. The parameter tuning after 50 iterations has been performed whereby weights of the various losses for the generator are changed; more precisely, the weights for binary adversarial is reduced in the overall loss calculation. This is to handle the problem of discriminator becoming too strict and not letting the generator reconstruct better images. \\ 

\section{Results}

\begin{figure}[!htb]
  \centering
  \includegraphics[scale=.4]{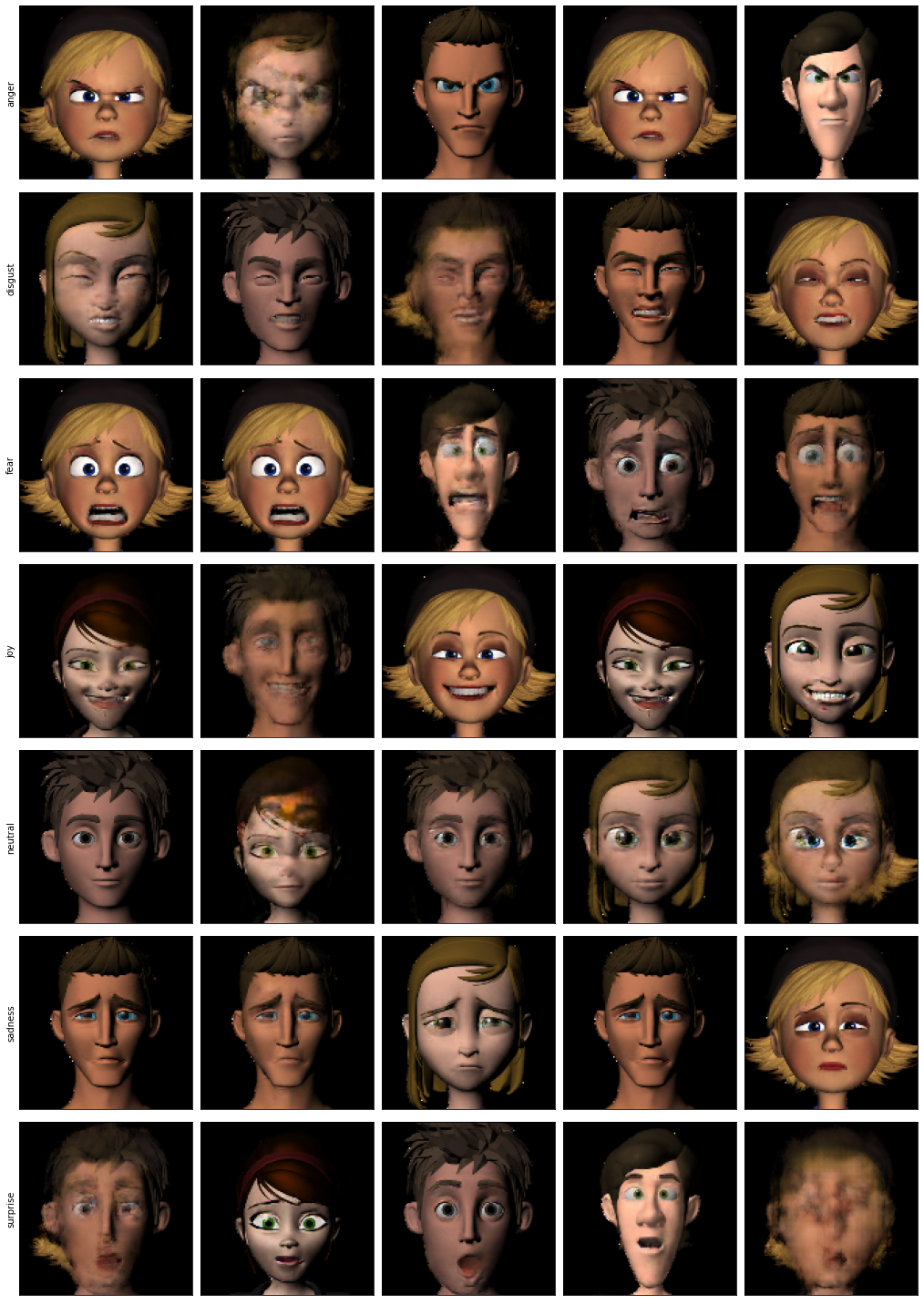}
  \caption{Images generated using Random Sampling in the Latent Space}
  \label{fig:fig4}
\end{figure}

One great thing about GAN architecture is that once the model has been trained, each part of the system (i.e. encoder, decoder, and discriminator) can be run separately. The decoder can be fed a random vector from the normal distribution, along with a desired affect vector and it generates an image of a cartoon character with the desired affect accordingly. The result of the decoder with random inputs can be visualized in the figure \ref{fig:fig4}. We can see that most of the randomly generated images are high quality and represent the desired facial expression very accurately. The variance among the generated images is also high which means the generator has accurately mapped the latent distribution of various characters in the dataset. \\

\begin{figure}[!htb]
  \centering
  \includegraphics[scale=.27]{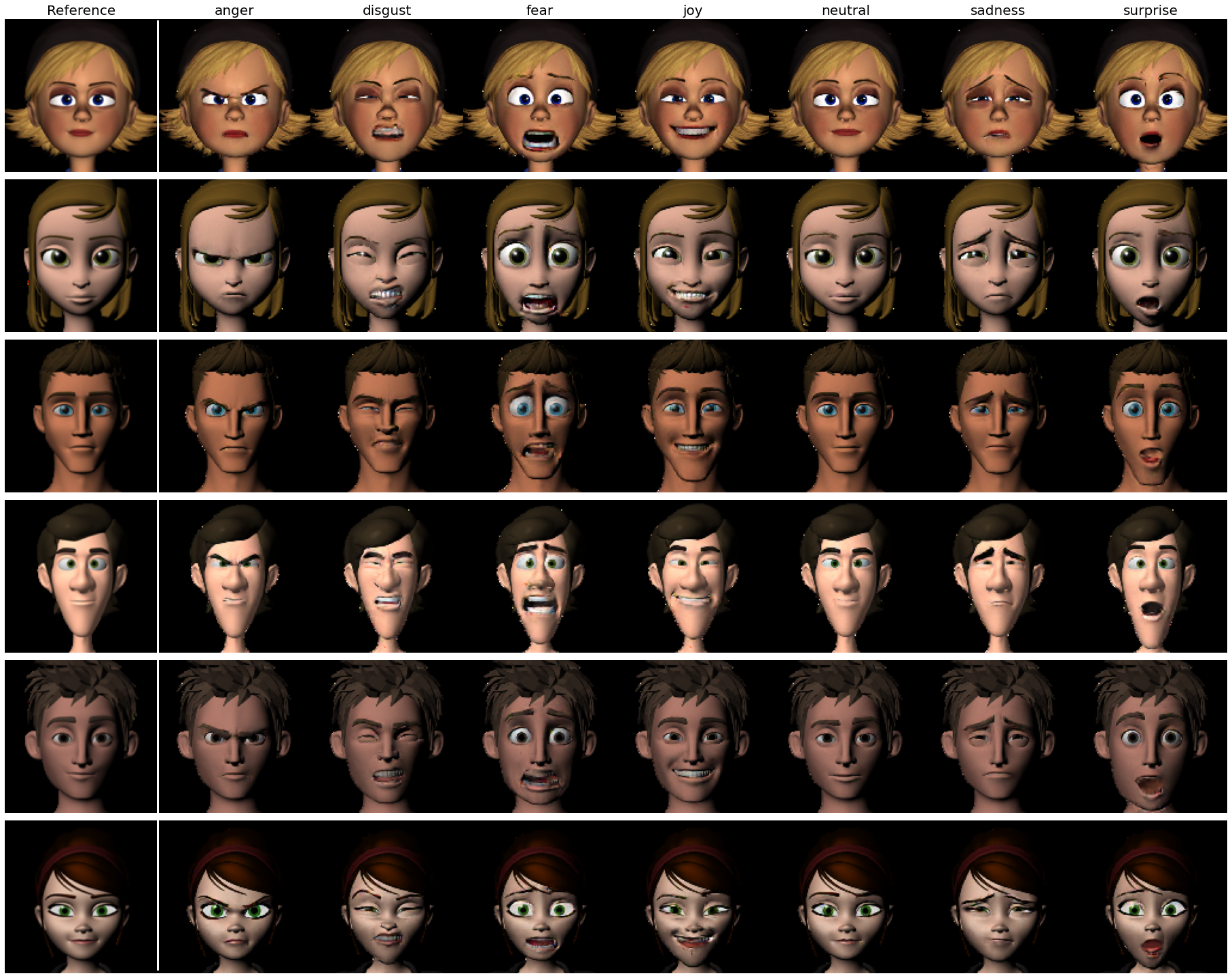}
  \caption{Results of various affect transformations for each character identity}
  \label{fig:fig5}
\end{figure}

In order to further analyze the results, we take a set of neutral images of each character as a reference image and then apply the encoder on them, thus obtaining a set of latent vectors as a result. These latent vectors along with a set of desired facial expressions, one for each affect, is fed into decoder network. The output can be seen in the figure \ref{fig:fig5}. We can see that each reference image with neutral expression has been accurately transformed into another desired expression. \\

\begin{figure}[!htb]
  \centering
  \includegraphics[scale=.3]{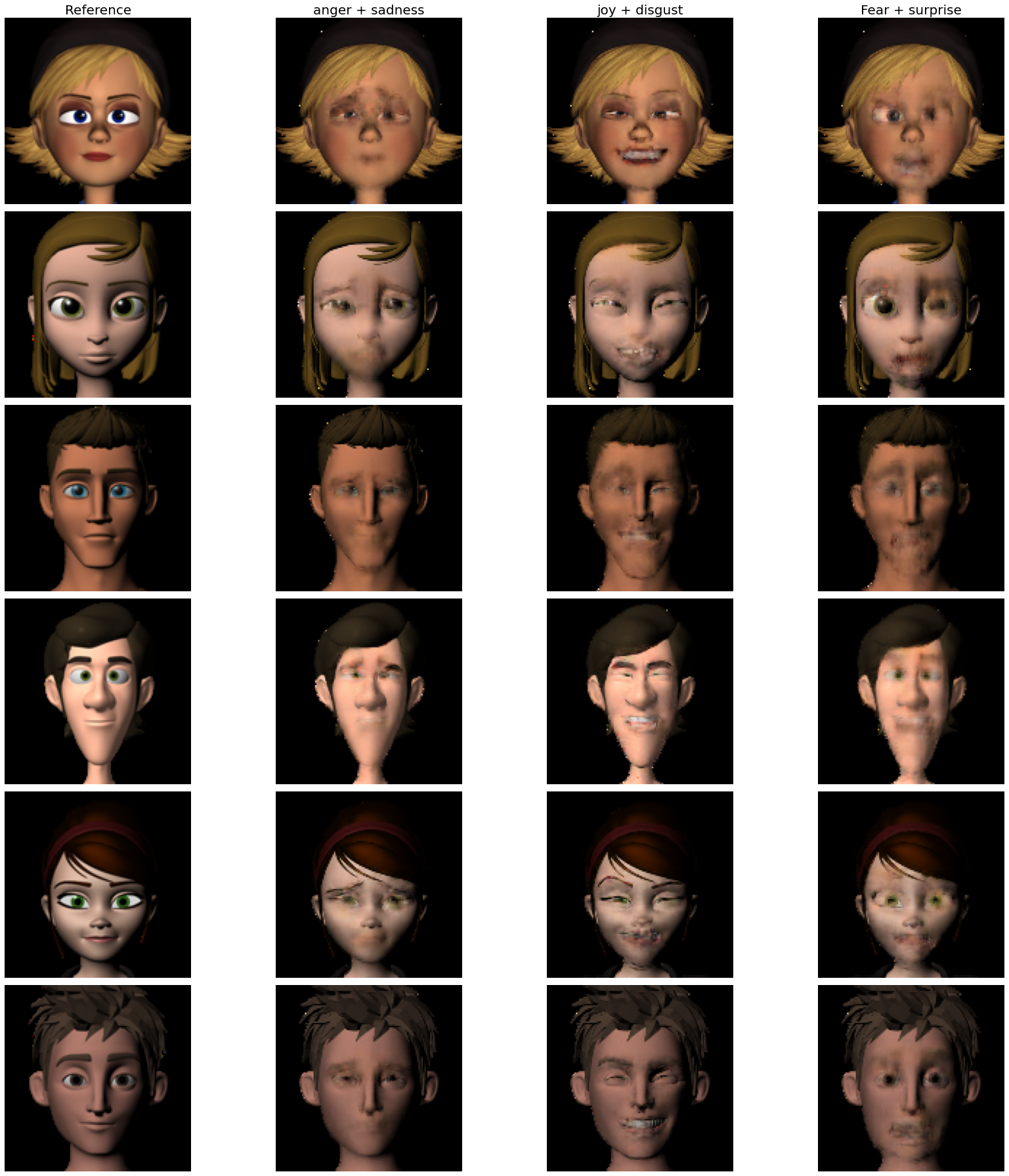}
  \caption{Image Generation for mixed Expressions}
  \label{fig:fig6}
\end{figure}

Since the model is robust at generating facial expression of a particular type, we can experiment further and try to generate complex expressions which can be a combination of multiple simple expressions. We accomplish this by first encoding a character with neutral facial expression into latent space and then supplying an affect condition such that it turns on two facial expressions at the same time. More specifically, we try to generate three complex expressions: anger+sadness, joy+disgust, and fear+surprise. The results of such an experiment can be seen in the figure \ref{fig:fig6}. Although not very crisp and clear results, unlike the simple expressions, we see that generator nevertheless, generates something similar to a set of complex emotions. Keeping in mind that the network was never trained on the complex expressions, it is a fairly good output. The results can further be improved by fine-tuning on respective complex expressions with a few shots at most. \\

The quantitative results of the training as well test set are also recorded and can be seen in \ref{tab:accuracy}. These are the results of classification accuracy obtained by discriminator network for real as well as the fake images. 

\label{sec:results}

\begin{table}
\label{tab:accuracy}
 \caption{Accuracy of the Classification Network}
  \centering
  \begin{tabular}{ccccc}
    \toprule

    & \multicolumn{2}{c}{Real}  & \multicolumn{2}{c}{Fake}                 \\
    \cmidrule(r){2-3}    
	\cmidrule(r){4-5}    
     & Binary     & Multi     & Binary  & Multi \\

    \midrule
    Train & 0.981 & 1.0  & 0.998  &  0.994 \\
	Val   & 0.913 & 0.997 & 1.0   &  0.995 \\
    
    \bottomrule
  \end{tabular}
\end{table}

\section{Conclusions}
In this article, we have proposed a robust facial expression generation generative adversarial network architecture. The proposed method has been evaluated on synthetic images of cartoon characters with a large variation of facial expressions exhibited by multiple characters. The results show that the architecture is robust in generation, classification and manipulation of the facial expressions. Further improvements should make it possible to transition to real images with a few shot learning. All in all the same method can be used as it is in the character animation applications. 

\clearpage
\bibliographystyle{unsrt}  
\bibliography{references}  

\end{document}